\documentclass[]{svproc}

\usepackage{url}

\usepackage{xcolor}
\usepackage{graphicx}
\usepackage{subcaption}
\usepackage[hidelinks]{hyperref}
\usepackage{cite}
\usepackage{overpic}
\usepackage{dsfont}
\usepackage{amsmath}
\usepackage{amssymb}
\usepackage{siunitx}
\usepackage[symbol]{footmisc}

\begin{document}
\mainmatter              
\title{Optimizing Force Signals from Human Demonstrations of In-Contact Motions}
\titlerunning{Optimizing Force Signals from Human Demonstrations}  
%
\author{Johannes Hartwig\footnote[2]{These authors contributed equally.} \and Fabian Viessmann\footnotemark[2] \and Dominik Henrich}
\authorrunning{Johannes Hartwig, Fabian Viessmann, Dominik Henrich} 
%
\tocauthor{Johannes Hartwig, Fabian Viessmann, Dominik Henrich}
\institute{Chair for Applied Computer Science III (Robotics and Embedded Systems), University of Bayreuth, D-95440 Bayreuth, Germany\\
\email{\{johannes.hartwig|fabian.viessmann\}@uni-bayreuth.de}}

\maketitle              

\begin{abstract}
	For non-robot-programming experts, kinesthetic guiding can be an intuitive input method, as robot programming of in-contact tasks is becoming more prominent.
	However, imprecise and noisy input signals from human demonstrations pose problems when reproducing motions directly or using the signal as input for machine learning methods.
	This paper explores optimizing force signals to correspond better to the human intention of the demonstrated signal. 
	We compare different signal filtering methods and propose a peak detection method for dealing with first-contact deviations in the signal.
	The evaluation of these methods considers a specialized error criterion between the input and the human-intended signal. 
	In addition, we analyze the critical parameters' influence on the filtering methods.
	The quality for an individual motion could be increased by up to \SI{20}{\percent} concerning the error criterion.
	The proposed contribution can improve the usability of robot programming and the interaction between humans and robots.
	
\end{abstract}

\section{Introduction and Related Work}
	Due to current market demands, raising the flexibility and agility in automation using cobots is an increasing trend in robotics \cite{Dietz2012}. 
	As small and medium-sized enterprises (SMEs) lack robot programming experts for economic reasons, enabling the available task experts to program and reconfigure robot systems easily is critical \cite{Riedelbauch2019, Perzylo2019}. 
	This allows them to contribute their task knowledge directly. Therefore, a future-oriented goal is to improve the usability of robot programming and the interaction between humans and robots \cite{Kildal2018}. 
	Utilizing the programming by demonstration (PbD) paradigm, specifically with kinesthetic guiding, can enable humans to program robot trajectories without prior robot programming knowledge \cite{Villani2018}. 
	Thus, a human can program robot trajectories by guiding the robot directly and fulfilling a given task only with his domain knowledge. 
	
	Nevertheless, specific tasks can only be adequately addressed partially through trajectory programming, such as when a robot task consists of polishing \cite{Amanhoud2019}, engraving \cite{Koropouli2011}, or grinding complex shapes \cite{Jinno1995}. 
	Here, continuous contact between the tool and the object becomes imperative, and a method to specify the contact forces to be applied over time and space is needed. 
	There are possibilities to achieve this simultaneous input of position and force. For example, in assembly tasks, the contact forces can be measured by a force/torque sensor located underneath a mounting bracket \cite{Bargmann2021}, or the needed robot stiffness is estimated by an electromyographic wristband from the muscle activation of the user \cite{Yang2019}. 
	The most prominent approach we will also use here is an additional force/torque sensor attached between the user's hand-guiding contact point and the robot's tool \cite{Montebelli2015, Steinmetz2015, Conkey2019}. 
	In this way, we can measure the robot's trajectory and the forces exerted on the environment during the demonstration by hand guiding. 
	
\begin{figure}
		\centering
		\begin{subfigure}{.6\textwidth}
			\centering
			\includegraphics[height=3.5cm,width=\linewidth]{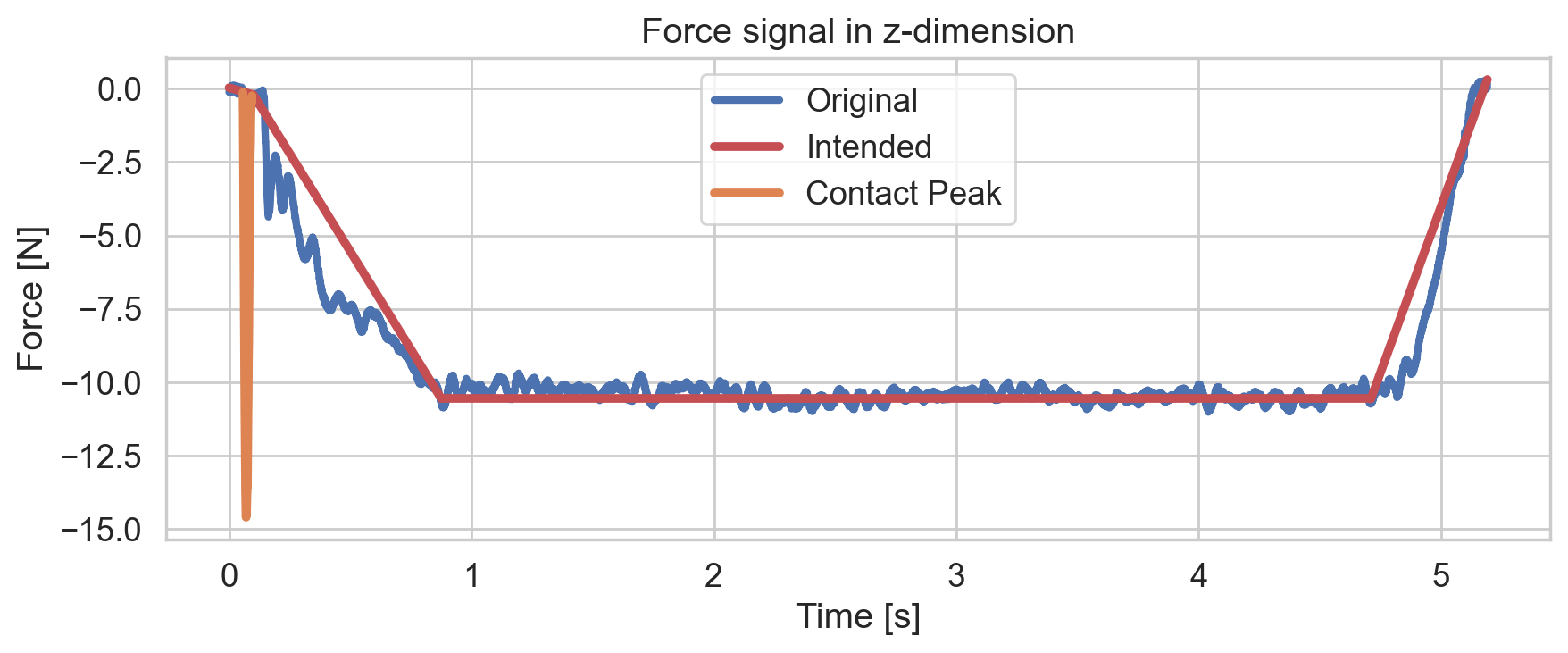}
			\caption{Exemplary signal}
			\label{fig:example_signal}
		\end{subfigure}%
		\begin{subfigure}{.4\textwidth}
			\centering
			\includegraphics[height=3.5cm,width=\linewidth]{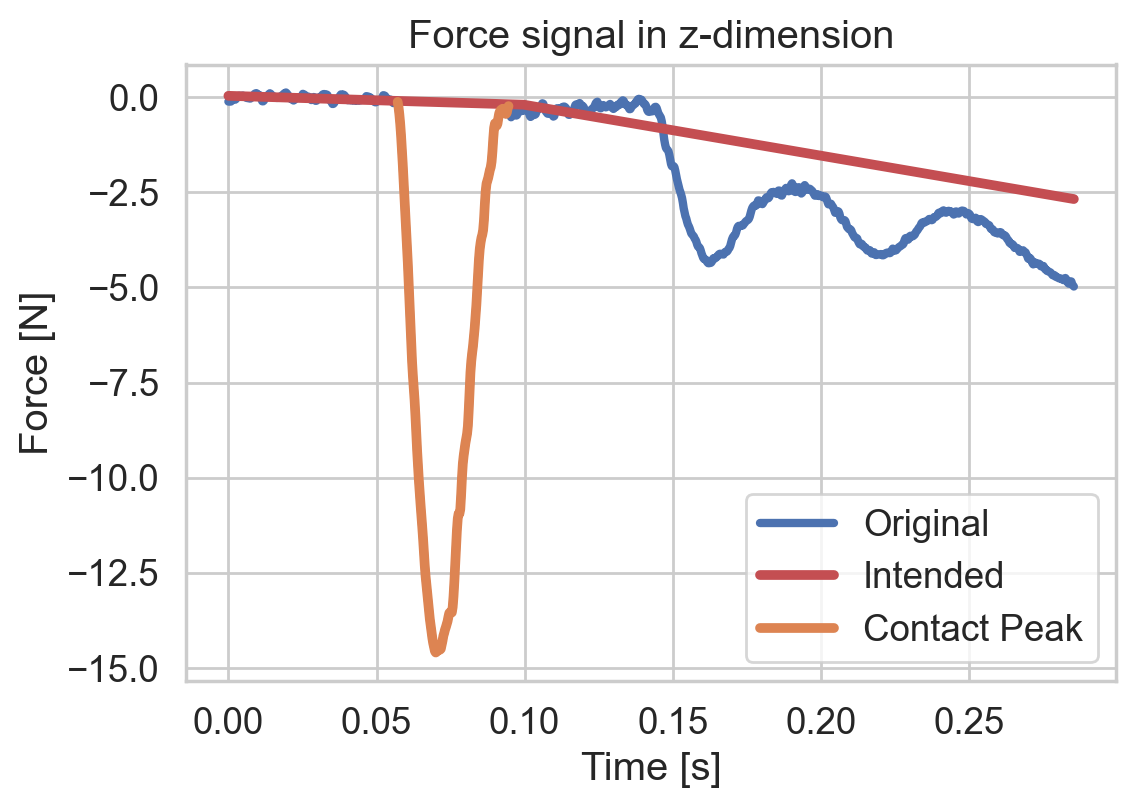}
			\caption{Exemplary contact peak}
			\label{fig:peak}
		\end{subfigure}
		\caption{Exemplary signal of a human demonstration (blue), an ideal signal (red) and a contact peak (orange) in Fig. \ref{fig:example_signal}. Contact peak is enlarged in Fig. \ref{fig:peak}.}
	\end{figure}

	However, the force profiles generated by humans during the demonstration are often noisy or inaccurate (see Fig. \ref{fig:example_signal}). 
	A prominent approach here is to learn from multiple demonstrations, often combining methods such as dynamic time warping (DTW), Gaussian mixture models (GMMs), or dynamical systems. 
	A categorizing overview for robot manipulation in contact can be found here \cite{Suomalainen2022}. 
	Otherwise, when programming the force profiles directly, these inaccuracies can be visualized and corrected by the user like in our previous work \cite{Hartwig2023a, Hartwig2023b}, or one can enhance them automatically using an error correction or signal filtering technique. 
	Most of these methods employ either frequency-domain filtering (e.g., using Fourier Analysis \cite{Cooley1965}), averaging over filter windows (e.g., Savitzky-Golay filter \cite{Savitzky1964}), or fit primitives (e.g., using RANSAC \cite{Fischler1981}). 
	Each method has different advantages and drawbacks. 
	We will explore these methods, applying their fundamental approach to our inaccurate signal.
	
	Therefore, this paper examines the question of to what extent measured force signals can be optimized so that they correspond better to the human intention of the demonstrated signal. 
	This can be used to minimize the error when reproducing the motion directly as in playback programming or as a signal pre-processing for machine learning methods to improve the signal-to-noise ratio. 
	We evaluate three different filtering techniques for force profiles and aim to contribute insights on what to consider in such a case for practitioners.
	
	First, Sec. \ref{sec:methodology} formalizes our input signal and error criterion before describing different signal filtering methods. 
	Subsequently, Sec. \ref{sec:evaluation} discusses their evaluation regarding the error criterion between the input and the human-intended signal. 
	The critical parameters of the filtering methods are especially considered here. 
	Sec. \ref{sec:conclusion} summarizes the paper and discusses future work. 
	
\section{Methodology}
\label{sec:methodology}
	When recording an in-contact motion through kinesthetic guiding, the robot configurations $\mathcal{Q} = (q_{1},\dots,q_{n})$ are recorded with $q_{i} \in \mathbb{R}^d$ and $d$ joints of the robot. 
	The motion's sampling frequency $f_S$ and the duration $T$ determines the number of demonstrated configurations by $n = f_S \cdot T$. 
	In addition, we record the force/torque value  $w_{i}\in \mathbb{R}^6$ for each $q_i$ using a force/torque sensor located between the user's hand-guiding contact point and the end-effector. This sensor measures the wrench $\mathcal{F}=(w_{1},\dots,w_{n})$ exerted by the user on the robot's tool as it interacts with the environment.
	Each $w_i$ contains three force and three torque values, one along each force/torque sensor coordinate axis.
	For this paper, we only consider one dimension at a time so that we can define a one-dimensional force/torque signal $\mathcal{F}_j=(w_{1,j},\dots,w_{n,j})$ with $w_{i,j}\in \mathbb{R}$ representing the force/torque value in dimension $j$ at time $i$. 
	We will refer to the signal $\mathcal{F}_j$ as $\mathcal{S} = (k_{1},\dots,k_{n})$.
	
	Our approach aims to optimize the force signals obtained from human demonstrations by kinesthetic guiding. 
	The goal is to minimize the error between a force signal $\mathcal{S}$ from a human demonstration and an ideal signal $\hat{\mathcal{S}} = (\hat{k}_{1},\dots,\hat{k}_{n})$ that meets the intentions of the user (see Fig. \ref{fig:example_signal}). 
	The user's intended signal is usually predefined by the given task.
	The deviations from the ideal signal occur due to various effects. 
	Firstly, there are inaccuracies due to the unfamiliar guidance of the tool via the robot arm. 
	Also, there are inaccuracies due to human muscle activation and sensor noise. 
	As no ideal force signal is usually available for a human demonstration, we define the ideal force signals for different scenarios (see Sec. \ref{sec:evaluation}). 
	We can define an optimization criterion based on these ideal signals and the recorded ones (see Sec. \ref{subsec:error}).
	
	To minimize the error criterion, we apply different filtering/smoothing algorithms to the force signal to generate an optimized signal $\tilde{S}$. 
	Each smoothing algorithm has a critical parameter that strongly influences the optimization result. 
	Therefore, we must find the critical parameter that minimizes the error criteria for each method. 
	Since force signals obtained from human demonstrations typically contain significant peaks due to the first contact with the environment, the force signals can not be used directly as input for the smoothing algorithms (see Fig. \ref{fig:example_signal}). 
	Large peaks in the signal distort the results. Thus, our approach consists of two steps. 
	The first step is identifying and removing unwanted peaks (see Sec. \ref{subsec:peak}). 
	The second step is applying different smoothing algorithms (see Sec. \ref{subsec:smoothing}): \textit{Sinc-in-time filter}, \textit{Savitzky-Golay filter}, and \textit{RANSAC line approximation}.

\subsection{Error Criterion}
\label{subsec:error}

	Regarding the error criterion, we first have to define how we want to weight the error components between the given signal $S$ and the optimal signal $\hat{S}$ at a given time $t$. 
	We can either use an L1 norm based on the absolute error value, like mean absolute error (MAE), or an L2 norm squaring the result, like mean squared error (MSE). 
	As we assume that outliers, i.e., a single sub-optimal demonstration by a user, should not be weighted more, we consider an L1 error criterion more suitable here.
	
	However, as seen in Fig. \ref{fig:example_signal}, the force signal of a human demonstration often deviates strongly from the expectation at the start and end. 
	In order to obtain information about the error in the central area of the signal, we need to weight the error at the start and end lower. 
	There is a similar line of thought in the error weighting of control loops. 
	Here, the error in the step response should not dominate the measure, and the oscillations later should also be considered. 
	
	For control loop problems, one of the prominent criterion functions is the integral of the time-multiplied absolute value of error (ITAE) criterion \cite{Joseph2022}.
	This criterion tries to solve the described problem by weighting the absolute error regarding the time $t$:
	\begin{align}
		J_{\text{ITAE}}= \int_{0}^{\infty} |e(t)| \cdot t\, dt
	\end{align}
	Since only the start is weighted less using the ITAE criterion, we modify it by using a bidirectional integral weighted from start and end for our finite signal of duration $T$:
	\begin{align}
		J_{\text{bidirectional-ITAE}} = \int_{0}^{\frac{T}{2}} |e(t)| \cdot t\, dt + \int_{\frac{T}{2}}^{T} |e(t)| \cdot (T-t)\, dt
	\end{align}
	We now adapt this criterion to our discrete case with the recorded signal $S$ and ideal $\hat{S}$:
	\begin{align}
		\tilde{J}(\mathcal{S},\hat{\mathcal{S}}) = \sum_{i=1}^{\lfloor \frac{n}{2} \rfloor} i \cdot |k_{i}-\hat{k}_{i}| + \sum_{i=\lfloor \frac{n}{2} \rfloor + 1}^{n}  (n + 1 - i) \cdot |k_{i}-\hat{k}_{i}|
	\end{align}
	In order to make signals of different lengths comparable, we also normalize the error regarding the sum of time weights:
	\begin{align}
		J(\mathcal{S},\hat{\mathcal{S}}) = \frac{ \sum_{i=1}^{\lfloor \frac{n}{2} \rfloor} i \cdot |k_{i}-\hat{k}_{i}| + \sum_{i=\lfloor \frac{n}{2} \rfloor + 1}^{n}  (n + 1 - i) \cdot |k_{i}-\hat{k}_{i}| }{\sum_{i=1}^{\lfloor \frac{n}{2} \rfloor} i + \sum_{i=\lfloor \frac{n}{2} \rfloor + 1}^{n} (n + 1 - i)} 
	\end{align}
	
	Although we normalize the criterion by the sum of time weights, only similar signals should be compared because a fair comparison requires the same number of significant changes (resp. step responses in the controller analogy). 
	In addition, the ratio of significant changes to oscillation phases should be similar. 
	Otherwise, they would be weighted differently using this criterion.
	This means only signals of similar individual motions should be compared using this criterion, not signals of entire demonstrated tasks.

\subsection{Contact Peak Removal}
\label{subsec:peak}

	When analyzing the signals from human demonstrations, we observe that, typically, force signals contain so-called contact peaks. 
	These represent a large deflection in the force signal, which is very short in time. 
	The peak occurs during initial contact between the tool and the environment, as humans typically establish contact with the environment too quickly and lose it again immediately. 
	Fig. \ref{fig:peak} depicts such a contact peak. 
	
	Therefore formally, a peak in our force signal $\mathcal{S}$ is defined as a sequence $\mathcal{P}=(k_{u},k_{u+1},\dots,k_{v-1},k_{v},k_{v+1},\dots,k_{w-1},k_{w})$ at times $i$ between peak start $u$ and end $w$ with $1\leq u < v < w \leq n$ and $\text{argmin}_{u\leq i \leq w} k_{i}=v$.
	Let now $\mathcal{S}'=(k'_{1},\dots,k'_{n})$ be the derivation of the force signal $\mathcal{S}$. 
	Since $\mathcal{S}$ is discrete, we must approximate the derivative using finite differences. 
	Then the following must hold: $k'_{i}<0$ with $u \leq i < v$, $k'_{v}=0$ and $k'_{i}>0$ with $v < i \leq w$. 
	
	Since this basic definition also applies to every force signal down- an upswing, we have to define further requirements:
	\begin{enumerate}
		\item When dealing with these unwanted short peaks, it must generally hold that the force $k_u$ at the end of the peak $u$ corresponds approximately to the force $k_w$ at the beginning of the peak $w$. 
		For this purpose, we use a relative threshold to set the difference in magnitude between the forces $k_u$ and $k_w$ to the total peak height $|k_u-k_v|$. 
		This relative threshold should be less than or equal to a given parameter $\delta$.
		\begin{align}
			\frac{|k_{u}-k_{w}|}{|k_{u}-k_{v}|} \leq \delta
		\end{align}
		\item When considering the derivative, there is a strong deflection for the decreasing and the increasing part of the contact peak. 
		This means the value changes significantly during the contact peak's time interval in relation to the rest of the signal. 
		Therefore, we can identify these as outliers utilizing the so-called z-score. 
		If the z-score deviates from the mean by more than $\mu$ standard deviations, we consider this value an outlier. 
		Therefore, let be $\bar{\mathcal{S}'}$ the mean value and $\sigma$ be the standard deviation of the signal $\mathcal{S}'$. Then the following must hold:
		\begin{align}
			\exists u\leq i < v:\frac{k'_{i}-\bar{\mathcal{S}'}}{\sigma} > \mu\land 
			\exists v < l \leq	w:\frac{k'_{l}-\bar{\mathcal{S}'}}{\sigma} > \mu
		\end{align}
		\item After the contact peak, additional smaller peaks may occur in the signal during the human step response. 
		However, we only intend to filter the first peak as it represents the beginning of a motion.
		Hence, we must take into account the temporal distance between potential peaks. 
		Let $\mathcal{P}_1$ and $\mathcal{P}_2$ be two peaks, Then the following must hold:
		\begin{align}
			u_2-w_1 \geq \tau
		\end{align}
	\end{enumerate}
	After using these conditions to identify all contact peaks, we remove the found peaks by setting $k_{i}=k_{u}$ for all $i$ in the peak interval $[u,w]$.

\subsection{Application of Filtering Techniques}
\label{subsec:smoothing}
	
\begin{figure}	
		\centering
		\includegraphics[width=.95\linewidth]{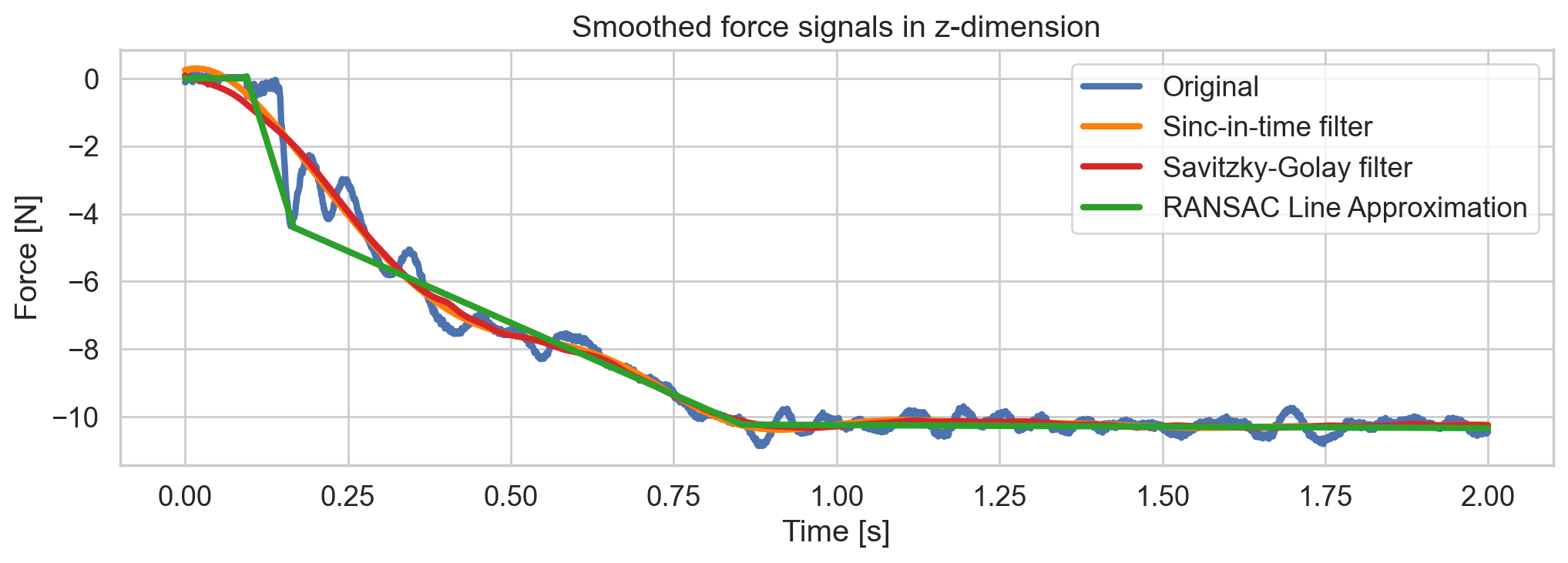}
		\caption{Application of the filtering techniques of an exemplary signal. Used parameters are $f_c=\SI{3}{\hertz}$ for the \textit{Sinc-in-time filter}, $h=\SI{0.25}{\second}$ for the \textit{Savitzky-Golay filter} and $l=\SI{1}{\second}$ for \textit{RANSAC line approximation}.}
		\label{fig:smoothing}
	\end{figure}

	After identifying and removing the signal's contact peaks, we can apply different filtering approaches: 
	Sinc-in-time filter using Fourier Transform, Savitzky-Golay filter, and RANSAC line approximation. 
	Each algorithm has a critical parameter that strongly influences the smoothing result. 
	These three algorithms are considered because each represents the basic idea of either frequency-domain filtering, averaging over a filter window, or fitting primitives.
	We want to consider all of these approaches for our optimization problem. 
	For an example of the application and the smoothed results, see Fig. \ref{fig:smoothing}.
	
	\paragraph*{Sinc-in-time filter:}
	For this, we determine the frequency components of the force signal first. 
	Since our signal is discrete, we use the Fast Fourier Transform \cite{Cooley1965}. 
	To smooth our force signal, we remove all frequencies above a predetermined cutoff frequency $f_c$ and obtain a filtered signal by executing the Inverse Fourier Transform. 
	Therefore, this is an optimal low-pass filter, and the critical parameter of this algorithm is the cutoff frequency $f_c$. 
	The lower the cutoff frequency is, the greater the smoothing, and vice versa.
	
	\paragraph*{Savitzky-Golay filter \cite{Savitzky1964}:} 
	This filter performs a weighted moving average using coefficients to perform a polynomial regression in a moving window with a low-degree polynomial. 
	The polynomial is then evaluated in the center of the window, and its value is used as the new signal value at this position.
	Therefore, the critical parameter of the Savitzky-Golay filter is the window width $h$; 
	The wider the window is, the greater the smoothing, and vice versa.
	
	\paragraph*{Random Sample Consensus (RANSAC) \cite{Fischler1981} Line Approximation:} 
	Here, we fit a model into the signal. 
	The RANSAC method first selects the number of data points required to instantiate the model to be fitted.
	The supporting data points of this model are then counted, and the model with the most supporters is selected. 
	The model is then fitted according to the least squares among these supporters so that outliers do not distort the fitting process. 
	In this work, we fit linear models into the force signals.
	Therefore, we have to execute the RANSAC algorithm multiple times to find multiple linear model approximations explaining the whole signal.
	There is an additional parameter to determine the granularity of the linear models, which is the maximum segment length $l$. 
	This length limits the time interval in which the respective model applies. 
	Therefore, the critical parameter of this smoothing algorithm is the maximum segment length $l$;
	The greater the length $l$ is, the greater the smoothing, and vice versa.
	
\section{Evaluation}
\label{sec:evaluation}

	To determine the critical parameters of the filtering techniques and subsequently evaluate the signal optimization, we gathered demonstrations by $10$ different users. 
	The participants were tasked to demonstrate $10$ different in-contact motions by kinesthetically guiding a Franka Emika Panda robot with an attached Schunk Gamma 6-axis force/torque sensor (z-axis measuring range $\pm \SI{100}{\newton}$, resolution $\SI{0.013}{\newton}$). 
	After getting familiar with the robot's hand guidance, the participants repeated every motion five times. 
	We provided them with visual feedback on the current force value using a bar representation of the current deviation and intuition about the motion (e.g., pressing on a wood dowel, or applying and pressing on an adhesive bond). A 3D-printed stump was attached to the robot's gripper with which the forces were to be exerted.
	
	The demonstrated motions can be divided into the following categories:
	Firstly, point with constant force in z-direction with $\SI{7.5}{\newton}$ and $\SI{15}{\newton}$, or linearly increasing force with $\SI{7.5}{\newton}$ and $\SI{15}{\newton}$; 
	Secondly, a part-wise linear path along a triangle with constant force in z-direction with $\SI{0}{\newton}$, $\SI{7.5}{\newton}$ and $\SI{15}{\newton}$; 
	Lastly, a circular path with constant force in z-direction with $\SI{0}{\newton}$, $\SI{7.5}{\newton}$ and $\SI{15}{\newton}$.

\subsection{Contact Peak Detection}
\label{subsec:eval_peak}

\begin{figure}	
		\centering
		\includegraphics[width=.9\linewidth]{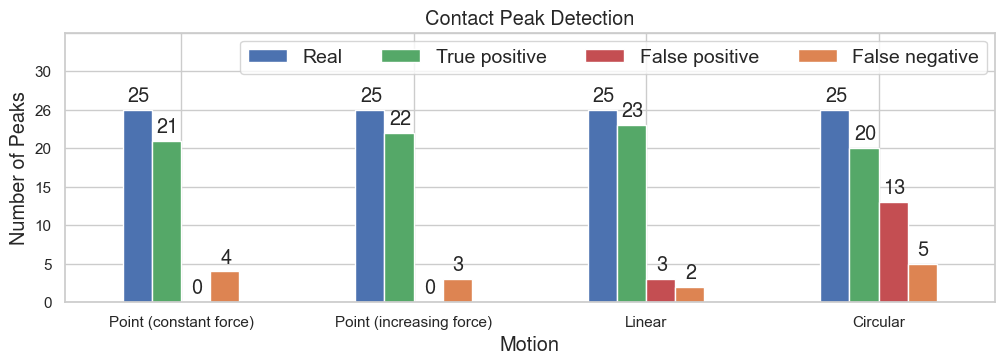}
		\caption{Evaluation of the automatically detected contact peaks compared to the manually labeled ones.}
		\label{fig:eval_peaks}
	\end{figure}

	To evaluate the peak detection (see Sec. \ref{subsec:peak}), we use 25 randomly selected demonstrations of each motion group. 
	We further split the point demonstrations into subgroups with constant or linear increasing force. 
	This differentiation makes sense here as the needed jump responses demonstrated by the users differ significantly, which may influence the peak detection results. 
	All contact peaks were labeled manually for the evaluation and then compared with the automatically detected peaks. 
	We distinguish between correctly recognized peaks (true positives), unrecognized peaks (false negatives), and additional wrongly recognized peaks (false positives). 
	Peak detection was performed with empirically determined parameters $\delta = 0.15$, $\mu = 3$, and $\tau = 600$.  
	Fig. \ref{fig:eval_peaks} shows the results.
	
	When analyzing the results of the point demonstrations with a constant force, it is evident that \SI{84}{\percent} of the peaks are correctly identified.
	With the chosen parameter set, peak detection works well for point demonstrations, as the results with increasing force were similar. For the linear motions, \SI{92}{\percent} of the peaks are correctly detected.
	However, three false positive peaks originating from the signal's noise are detected.
	For circular motions, slightly fewer peaks are correctly detected at \SI{80}{\percent}. 
	A total of $13$ false positive peaks are identified. 
	Here, whether these peaks belong in the false positive class is unclear. 
	We observed that it is challenging to exert a force while the user has to move the robot. 
	The robot may lose contact with the environment during the demonstration, representing a contact peak when it regains contact with the environment. 
	Also, the force applied may jump through stick-slip effects.
	
	Overall, peak detection gives good results, with an average sensitivity of \SI{86}{\percent}. 
	However, false positives cannot be ruled out, especially when the human demonstrator simultaneously applies force and moves the robot.

\subsection{Filtering Techniques}
\label{subsec:eval_smoothing_algorithms}
	
\begin{figure} [t]
		\centering
		\begin{subfigure}{.33\textwidth}
			\centering
			\includegraphics[width=\linewidth]{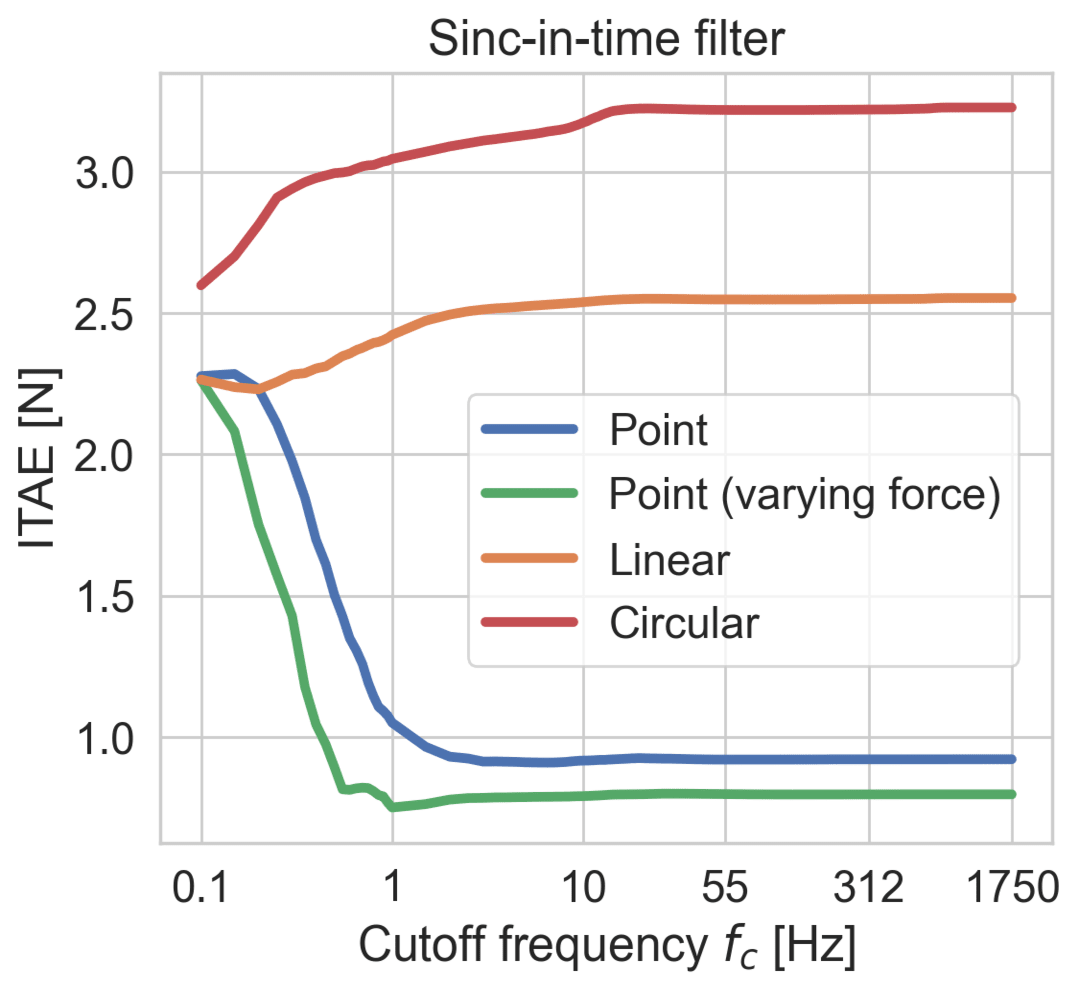}
			\caption{Sinc-in-time filter}
			\label{fig:ft_err}
		\end{subfigure}%
		\begin{subfigure}{.33\textwidth}
			\centering
			\includegraphics[width=\linewidth]{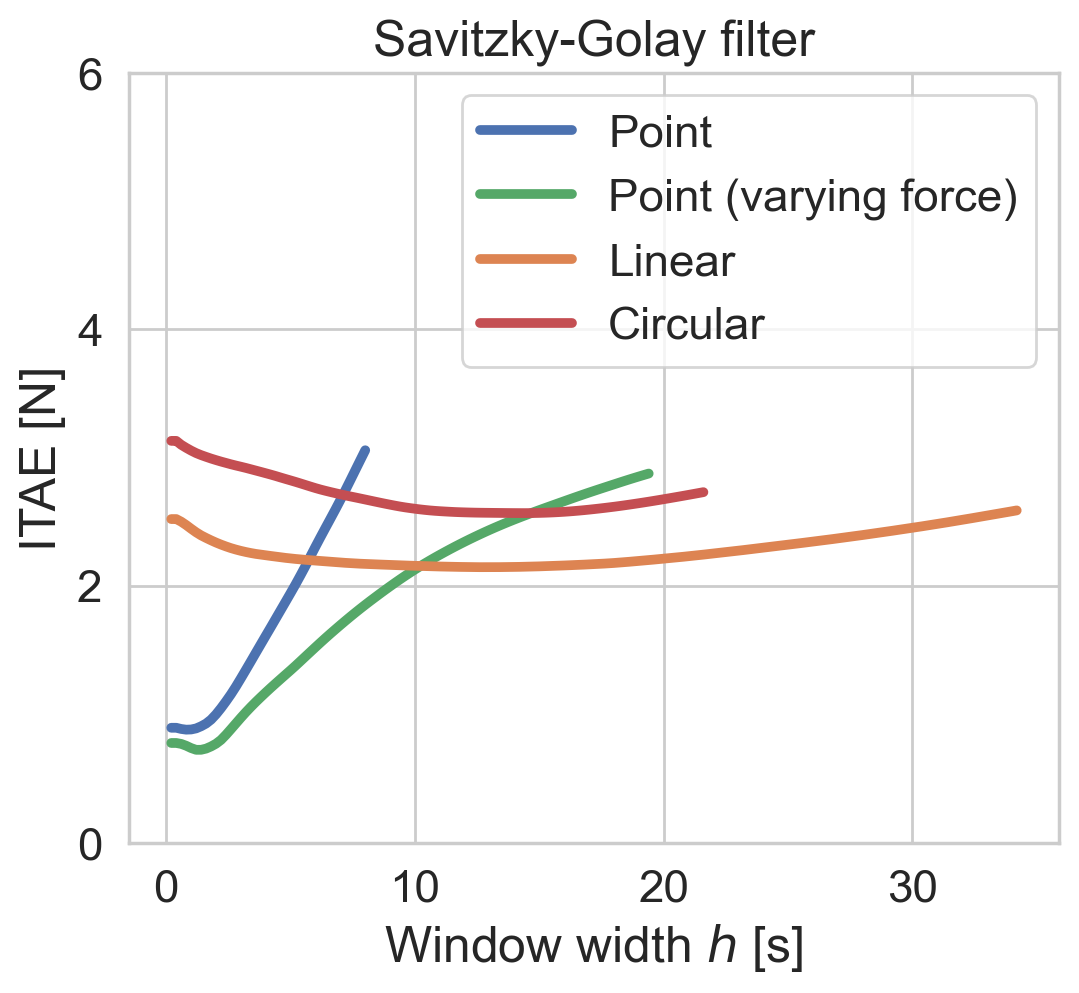}
			\caption{Savitzky-Golay filter}
			\label{fig:sg_err}
		\end{subfigure}
		\begin{subfigure}{.33\textwidth}
			\centering
			\includegraphics[width=\linewidth]{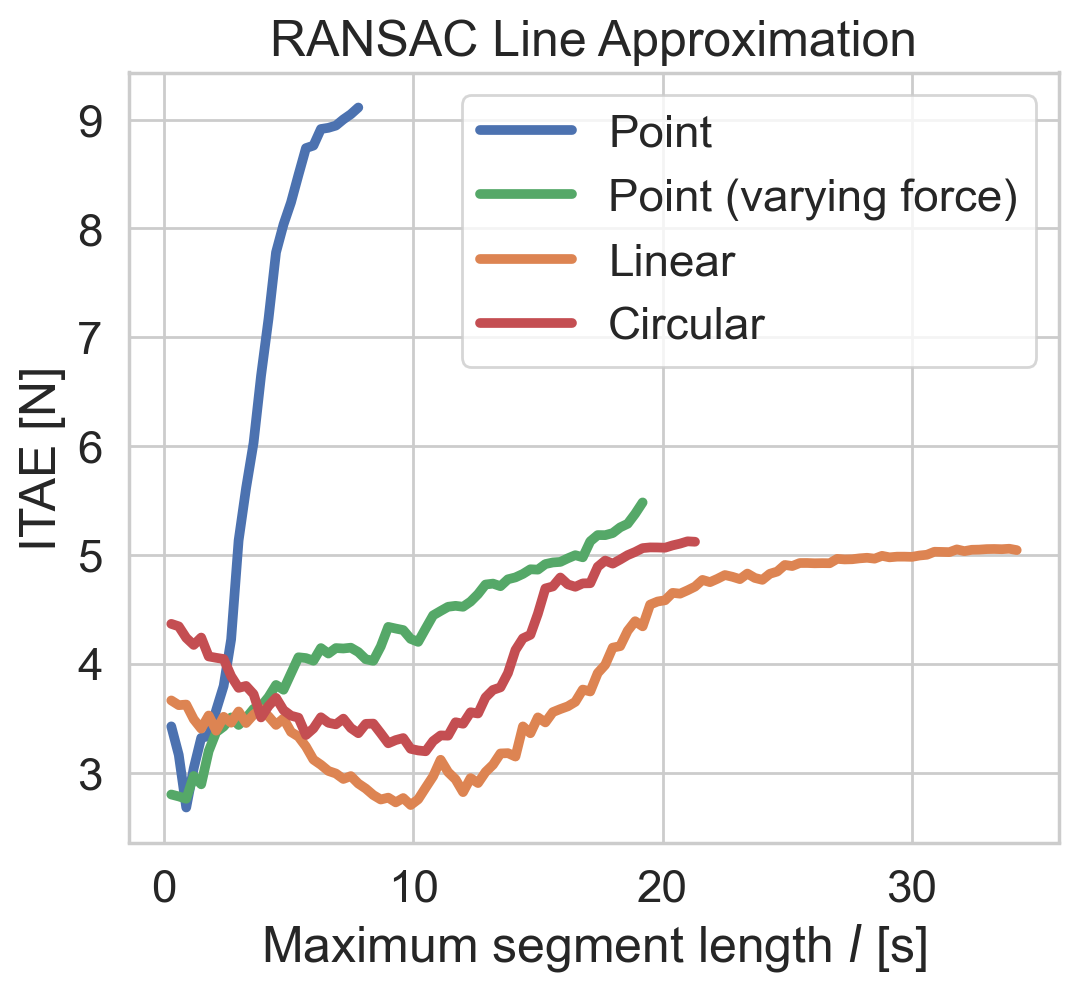}
			\caption{RANSAC}
			\label{fig:ransac_err}
		\end{subfigure}
		\caption{The error criterion averaged over all motions of each category plotted against the critical parameter of each method.}
		\label{fig:eval_smoothing}
	\end{figure}

	As a final step, we have to determine the critical parameter for each filtering technique that minimizes the error to the optimal motion on average for all demonstrations. 
	To achieve this, we sampled the parameter space for each approach and smoothed each signal with the sampled critical parameter. 
	The individual error curves are then averaged (Fig. \ref{fig:eval_smoothing}). 
	In addition to the mentioned demonstrations, we also consider point demonstrations with varying force profiles (step- and parabola-shaped) to analyze the influence of force complexity on optimization.
	These were generated with a different user group.
	
	With the Sinc-in-time filter, the minimum for the point demonstration is $f_c=\SI{6.5}{\hertz}$, and for the point demonstration with varying force, it is $\SI{1.0}{\hertz}$.
	Accordingly, the difficulty of the force profile to be applied influences the smoothing, and the more complex the force profile, the greater the smoothing must be. 
	In comparison, the cutoff frequency for linear demonstrations is significantly lower and is $f_c=\SI{0.15}{\hertz}$. 
	For circular demonstrations, this is even $f_c=\SI{0.1}{\hertz}$.
	This leads to the conclusion that the smoothing must be increased if the human moves the robot in addition to exerting force. 
	Furthermore, the minimum error for linear and circular demonstrations is still much higher than for point demonstrations.
	From this, we conclude that the difficulty of the positional motion should be weighted more heavily than the difficulty of  force profile.
	
	We observe similar results for the Savitzky-Golay filter and RANSAC line approximation (point:  $h=\SI{0.8}{\second}$, $l=\SI{0.9}{\second}$; point with varying force: $h=\SI{1.2}{\second}$, $l=\SI{0.9}{\second}$; linear:  $h=\SI{12.8}{\second}$, $l=\SI{9.9}{\second}$; circular:  $h=\SI{14.4}{\second}$, $l=\SI{10.5}{\second}$).
	When comparing the relative improvement between the input signal and filtered signal for linear demonstrations, the Sinc-in-time filter (\SI{18.47}{\percent}) performs similarly to the Savitzky-Golay filter (\SI{15.96}{\percent}). 
	RANSAC line approximation (\SI{-5.82}{\percent}) results in a worsening due to the inaccurate approximation of the signal by lines. 
	Therefore, a piece-wise linear approximation with our line model using RANSAC is not useful considering the given input signals, although many of our predefined test motions are linear.
	Overall, we can improve the error measure with Sinc-in-time and Savitzky-Golay filters, but the best parameter values differ heavily (see Fig. \ref{fig:eval_smoothing}). 
	This becomes particularly clear if, for example, we use the Savitzky-Golay filter with the point motion parameter for the circular motion (\SI{4.69}{\percent}) instead of the best parameter for the circular motion (\SI{20.45}{\percent}). 
	Thus, additional information about the motion must be considered, or user assistance is required.
	
\section{Conclusion}
\label{sec:conclusion}

	This paper compares different methods to optimize force signals from kinesthetic guiding. 
	We proposed a method to remove contact peaks in force signals and showed that it is possible to filter signals so that the smoothed signal corresponds more closely to the human intention. 
	Two different approaches improve the error measure and might be helpful for varying use cases. 
	Although significant improvement is achieved, a high degree of filtering is required, particularly for motions in which the human demonstrator exerts a force and moves the robot simultaneously.
	This phenomenon is mainly observed in the different critical parameters for the motions in Sec. \ref{subsec:eval_smoothing_algorithms}.
	Therefore, future work should address the modeling of critical parameters dependent on the positional motion.
	In addition, it should be investigated to what extent such filtering offers advantages when learning robot motions utilizing machine learning methods.

%


\end{document}